\documentclass[twoside,11pt,a4paper]{article}
\usepackage{arxiv,amsfonts,amsmath,tikz,bm,hyperref,microtype,pgfplots,graphicx}
\usetikzlibrary{patterns,arrows,calc,positioning} 

\jmlrheading{}{}{}{}{}{Lopez-Paz, Muandet and Recht}
\ShortHeadings{The Randomized Causation Coefficient}{Lopez-Paz, Muandet and Recht}

\definecolor{mydarkblue}{rgb}{0,0.08,0.45}

\hypersetup{
  colorlinks = true,
  linkcolor  = mydarkblue,
  citecolor  = mydarkblue,
  filecolor  = mydarkblue,
  urlcolor   = mydarkblue
}

\begin{document} 

\title{The Randomized Causation Coefficient}

\author{\name David Lopez-Paz\thanks{This project was conceived while DLP was
visiting BR at University of California, Berkeley.} \email david@lopezpaz.org
\\
        \addr Max-Planck Institute for Intelligent Systems\\
        \addr University of Cambridge\\
        \name Krikamol Muandet \email krikamol@tuebingen.mpg.de \\
        \addr Max-Planck Institute for Intelligent Systems\\
        \name Benjamin Recht  \email brecht@berkeley.edu\\
        \addr University of California, Berkeley\\
}

\editor{TBD}
\maketitle

\begin{abstract}%
  We are interested in learning causal relationships between pairs of random
  variables, purely from observational data. To effectively address this task,
  the state-of-the-art relies on strong assumptions regarding the mechanisms
  mapping causes to effects, such as invertibility or the existence of additive
  noise, which only hold in limited situations.
  On the contrary, this short paper proposes to \emph{learn} how to perform
  causal inference directly from data, and without the need of feature
  engineering. In particular, we pose causality as a kernel mean embedding
  classification problem, where inputs are samples from arbitrary probability
  distributions on pairs of random variables, and labels are types of causal
  relationships.
  We validate the performance of our method on synthetic and real-world data
  against the state-of-the-art. Moreover, we submitted our algorithm to the
  ChaLearn's ``Fast Causation Coefficient Challenge'' competition, with which
  we won the fastest code prize and ranked third in the overall leaderboard.
\end{abstract}
\vskip 1.5 cm

\section{Introduction}\label{sec:intro}
According to Reichenbach's common cause principle \citep{Reichenbach56:Time},
the dependence between two random variables $X$ and $Y$ implies that either $X$
causes $Y$ (denoted by $X \rightarrow Y$), or that $Y$ causes $X$ (denoted by
$Y \rightarrow X$), or that $X$ and $Y$ have a common cause. In this note, we
are interested in distinguishing between these three possibilities by using
samples drawn from the joint probability distribution $P(X,Y)$.

Two of the most successful approaches to tackle this problem are the
information geometric causal inference method \citep{Daniusis12,Janzing14}, and
the additive noise model~\citep{Hoyer09,Peters14:ANM}. 

First, the Information Geometric Causal Inference (IGCI) is designed to infer
causal relationships between variables related by invertible, noiseless
relationships.  In particular, assume that there exists a pair of functions or
\emph{mapping mechanisms} $f$ and $g$ such that $Y = f(X)$ and $X = g(Y)$. The
IGCI method decides that $X \rightarrow Y$ if $\rho(P(X),|\log(f'(X))|) \ll
\rho(P(Y),|\log(g'(Y))|)$, where $\rho$ denotes Pearson's correlation coefficient.
IGCI decides $Y \rightarrow X$ if the opposite inequality holds, and abstains
otherwise.  The assumption here is that the cause random variable is
independently generated from the mapping mechanism; therefore it is unlikely to
find correlations between the density of the former and the slope of the latter. 

Second, the additive noise model (ANM) assumes that the effect variable is
equal to a nonlinear transformation of the cause variable plus some independent
random noise. In particular, let $Y = f(X) + N_X$ and $X = g(Y) + N_Y$, where
$N_X$ and $N_Y$ are the respective residuals unexplained by $f$ and $g$.  Then,
the model will conclude that $X \rightarrow Y$ if the pair of random variables
$(X, N_X)$ are independent but the pair $(Y, N_Y)$ is not.  The algorithm will
conclude $Y \rightarrow X$ if the opposite claim is true, and abstain
otherwise. The additive noise model has been extended to study post-nonlinear
models of the form $Y = h(g(X)+N_X)$, where $h$ a monotone function
\citep{Zhang09}. The consistency of causal inference under the additive noise
model was established by \citet{Kpotufe13}.

As it becomes apparent from the previous exposition, there is a lack for a
general method to infer causality without assuming knowledge about
the (non) existence of noise.  Moreover, it is desirable to be able to readily
extend inference to other new model hypotheses without incurring in the
development of a new, specific algorithm. Motivated by this issue, we raise the
question, \emph{Is it possible to automatically learn patterns revealing causal
relationships between random variables from large amounts of labeled data?}

\section{Learning to learn causal inference}

Unlike the methods described in the previous section, we propose a data-driven
approach to build a flexible causal inference engine.  To do so, we assume
access to some set of pairs $\mathcal{D} =\{(S_i,l_i)\}_{i=1}^n$, where the
sample $S_i =\{(x_{ij},y_{ij})\}_{j=1}^{n_i}$ is drawn from the joint
distribution of the two random variables $X_i$ and $Y_i$, which obey the causal
relationship denoted by the label $l_i$.  To simplify exposition,
consider the labels $l_i = 1$, denoting $X \rightarrow Y$, and $l_i = -1$,
standing for $Y \rightarrow X$.  Using the dataset $\mathcal{D}$, we build a
causal inference algorithm in two steps.  First, an $m$-dimensional vector of
features $\bm m_i$ is extracted from each sample $S_i$, to meaningfully
represent the corresponding distribution $P_i(X_i,Y_i)$. Second, we use the set
$\{\bm m_i, l_i\}_{i=1}^{n}$ to train a binary classifier, later used to
predict the causal relationship between new, unseen pairs of random variables.
This framework can be straightforwardly extended to also infer the ``common
cause'' and ``independence'' cases, by introducing two extra labels. 

Our setup is fundamentally different from the standard classification problem
in the sense that the inputs to the learners are samples from probability
distributions, rather than real-valued vectors of features
\citep{Muandet12:SMM,Szabo14}.  In particular, we make use of two assumptions.
First, we assume the existence of a \emph{mother distribution}
$\mathcal{M}(\mathcal{P},\{-1,+1\})$ from which all paired probability
distributions $P_i(X_i,Y_i) \in \mathcal{P}$ and causal labels $l_i \in
\{-1,+1\}$ are sampled, where $\mathcal{P}$ is the set of all distributions on
two scalar random variables. Second, we speculate\footnote{It is widely agreed
that no absolutely general causal inference is possible
\citep{Cartwright94:Nature}.} that the causal relationships $l_i$ can be
inferred in most cases from observable properties of the distributions $P_i$.
While these assumptions do not hold in generality, our experimental evidence
suggests their wide applicability in real-world data.

The rest of this paper is organized as follows.  Section~\ref{sec:features}
elaborates on how to extract the $m-$dimensional feature vectors $\bm m_i$ from
each causal sample $S_i$.  Section~\ref{sec:exps} provides empirical evidence
to validate our methods. Section~\ref{sec:future} closes the
exposition by commenting on future research directions. 

\section{Featurizing distributions with kernel mean embeddings}
\label{sec:features}
Let $P$ be the probability distribution of some random variable $X$ taking
values in $\mathbb{R}^d$.  Then, the \emph{kernel mean embedding} of $P$
associated with the positive definite kernel function $k$ is 
\begin{equation}
  \label{eq:meank}
  \mu_k(P) := \int_{\mathbb{R}^d} k(\bm x, \cdot) \mathrm{d}P(\bm x) \in
  \mathcal{H}_k,
\end{equation}
where $\mathcal{H}_k$ is the reproducing kernel Hilbert space associated with
the kernel $k$ \citep{Berlinet04:RKHS,Smola07Hilbert}. The most attractive
property of $\mu_k$ is that it uniquely determines each distribution $P$ when
$k$ is a characteristic kernel
\citep{Sriperumbudur10:Metrics}. In another words,
$\|\mu_k(P)-\mu_k(Q)\|_{\mathcal{H}_k}=0$ iff $P=Q$. Examples of characteristic
kernels include the popular squared-exponential
\begin{equation}\label{eq:gauss}
  k(\bm x, \bm x') = \exp\left(-\gamma \|\bm x- \bm x'\|_2^2\right), \text{ for }
  \gamma >0,
\end{equation}
which will be used throughout this work.

However, in practice, it is unrealistic to assume access to the true
distribution $P$, and consequently to the true embedding $\mu_k$. Instead, we
often have access to a sample $\bm X = (\bm x_1, \ldots, \bm x_n) \in
\mathbb{R}^{d \times n}$ drawn i.i.d.  from $P$. Then, we can approximate
\eqref{eq:meank} by 
\begin{equation}
  \label{eq:meank2} 
  \hat{\mu}_k(\bm X) := \frac{1}{n} \sum_{i=1}^n k(\bm x_i, \cdot) \in \mathcal{H}_k.
\end{equation}
Though it can be improved \citep{Muandet14:KMSE}, the estimator
\eqref{eq:meank2} is the most common due to its ease of implementation. We can view
\eqref{eq:meank} and \eqref{eq:meank2} as a feature representation of the
distribution $P$ and its approximation, respectively. 

For some kernels, the feature maps \eqref{eq:meank} and \eqref{eq:meank2} do
not have a closed form, or are infinite dimensional. This translates into the
need of kernel matrices, which require at least $O(n^2)$ computation. In order
to alleviate these burdens, we propose to compute a low-dimensional
approximation of \eqref{eq:meank2} using random Fourier features
\citep{Rahimi07}. In particular, if the kernel $k$ is shift-invariant, we can
exploit Bochner's theorem \citep{Rudin62} to construct a randomized
approximation of \eqref{eq:meank2}, with form
\begin{equation}\label{eq:meank3}
  \hat{\mu}_{k,m}(\bm X) = \frac{1}{n} \sum_{i=1}^n \left[ \cos(\bm w_1' \bm
  x_i + b_1), \ldots, \cos(\bm w_m' \bm x_i + b_m)\right]' \in
  \mathbb{R}^m,
\end{equation}
where the vectors $\bm w_1, \ldots, \bm w_m$ are sampled from the Fourier
transform of $k$, and $b_1, \ldots, b_m \sim \mathcal{U}(0, 2\pi)$. The
squared-exponential kernel in \eqref{eq:gauss} is shift-invariant, and can be
approximated in this fashion when setting $\bm w_i \sim \mathcal{N}(\bm 0,
2\gamma \bm I)$.  These features can be computed in $O(mn)$ time and stored in
$O(1)$ memory. Importantly, the low dimensional representation
$\hat{\mu}_{k,m}$ is amenable for the off-the-shelf use with any standard
learning algorithm, and not only kernel-based methods.

In the following, we denote the featurization of each causal sample $S_i =
\{(x_{ij},y_{ij})\}_{i=1}^{n_i}$ as 
\begin{equation}
  \label{eq:ourfeatz}
  \bm m_i :=
  \left[
  \hat{\mu}_{k,m/3}\left(\{x_{ij}\}_{j=1}^{n_i}\right),
  \hat{\mu}_{k,m/3}\left(\{y_{ij}\}_{j=1}^{n_i}\right),
  \hat{\mu}_{k,m/3}\left(\{(x_{ij},y_{ij})\}_{j=1}^{n_i}\right)
  \right],
\end{equation}
where we have used \eqref{eq:meank3} to embed the marginal distribution of the
random variable $X$, the marginal distribution of the random variable $Y$, and
the joint distribution of both.

Using the assumptions introduced in Section~\ref{sec:intro}, the data $\{\bm
m_i,l_i\}_{i=1}^n$ and a binary classifier, we can now pose causal inference as
a supervised learning problem.

\section{Numerical simulations}\label{sec:exps}
We present two experiments to validate the efficacy of the proposed data-driven
causal inference method, which we term the \emph{Randomized Causation
Coefficient}, or RCC. Throughout our experiments, we make use of the squared
exponential kernel $\eqref{eq:gauss}$ with $\gamma = 10$ to featurize each
sample into $300$ random projections as per \eqref{eq:ourfeatz}. The necessary
source-code to replicate this experiments is available at
\url{http://lopezpaz.org/code/rcc.tar.gz}.
\subsection{T\"ubingen data}
We generate a synthetic dataset $\mathcal{D} = \{S_i,l_i\}_{i=1}^{N}$, where
$S_i = \{x_{ij},y_{ij}\}_{j=1}^{n}$, $N = 5,000$, $n = 1,000$, and $l_1 =
\cdots = l_{N} = 1$. Each sample $S_i$ is constructed as follows:
\begin{enumerate}
  \item The \emph{cause} vector $(x_{ij})_{j=1}^{n}$ is sampled from a random
  GMM of $5$ components. The mixture weights are Uniformly distributed and
  normalized to sum to one. The mixture means and standard deviations are
  sampled from $\mathcal{N}(0,5)$, accepting only positive samples for the
  standard deviations. The cause vectors are then standarized.
  \item The \emph{noise} vector $(\epsilon_{ij})_{j=1}^{n}$ is sampled from a
  centered Gaussian with a variance sampled from the Uniform distribution.
  \item The \emph{mapping mechanism} is a smooth spline fitted using as
  input a $10$-dimensional grid from $\min((x_{ij})_{j=1}^n)$
  to $\max((x_{ij})_{j=1}^n)$, and as output $10$ Normally distributed
  knots.
  \item The \emph{effect} vector is computed as $y_{ij} =
  f_i(x_{ij})+\epsilon_{ij}$ for all $1 \leq j \leq n$, and then standarized.
\end{enumerate}

\begin{figure}[t]
  \begin{center}
    \includegraphics{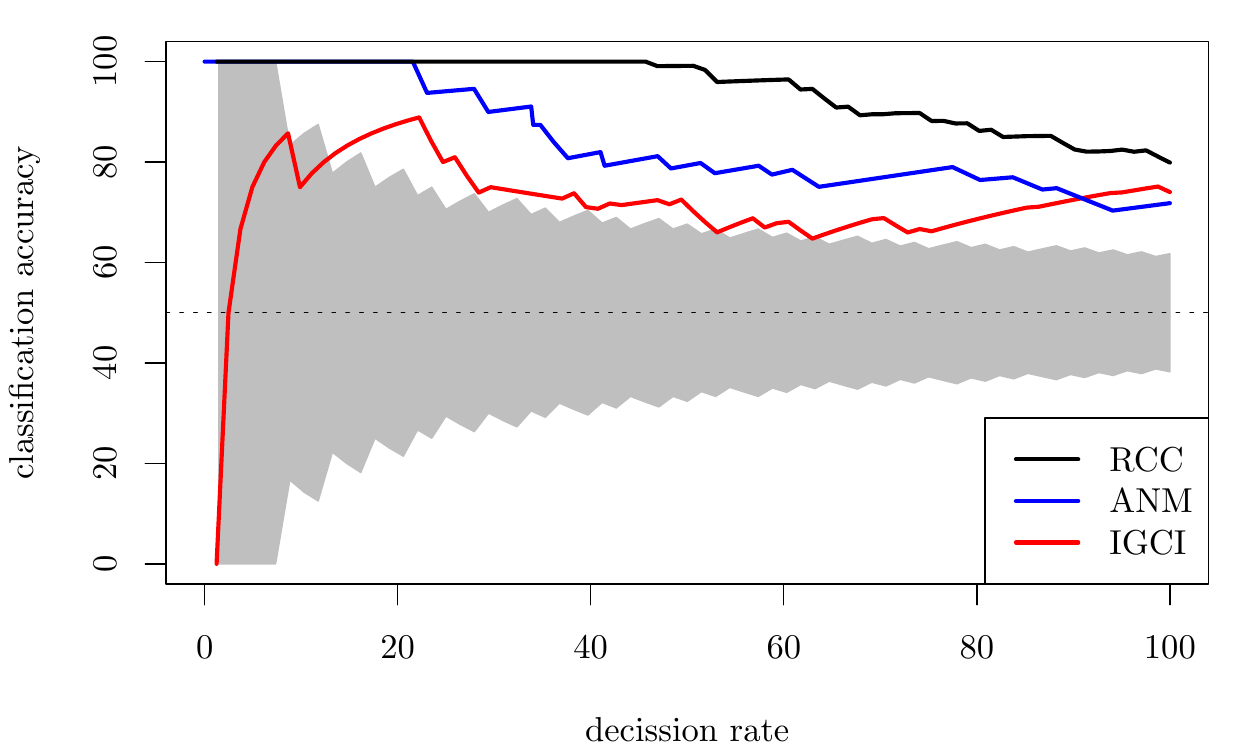} 
  \end{center}
  \vskip -0.5cm
  \caption{Accuracy of RCC, IGCI and ANM on the 82 scalar
  T\"ubingen pairs, as a function of decision rate. The grey area
  depicts accuracies not statistically significant.}
  \label{fig:tuebingen}
\end{figure}

Next, for each pair $(S_i, l_i)$, a new pair $(\{(y_{ij},x_{ij})\}_{j=1}^n,
-l_i)$ is generated and added to $\mathcal{D}$.  Finally, we train a random
forest of $500$ decision trees with a minimum of $1\%$ of samples per leaf on
the featurizations \eqref{eq:ourfeatz} of the dataset $\mathcal{D}$. We choose
to use a random forest in this experiment because we are interested in reducing
the variance of our predictions, since we are certain that the test data will
not exactly follow the same distribution as our synthetic, training data. We
evaluate the test classification accuracy of the random forest on the 82
one-dimensional T\"ubingen causal pairs \citep{Peters14:ANM,Tuebingen14}, which
are curated from heterogeneous real-world data. Figure \ref{fig:tuebingen}
plots the classification accuracy of RCC, IGCI and ANM versus the fraction of
decisions that the algorithms are forced to make out of the 82 pairs. To
compare these results to other lower-performance alternatives please refer to
\citep{Janzing12}. Overall, RCC shows the best performance, surpassing the
state-of-the-art with a classification accuracy of $79\%$ when forced to infer
the causal direction of all the 82 T\"ubingen pairs. The confidence of RCC is
computed as the output probabilities of the random forest.

Note that our training data generating process is incredibly basic, and that a
more powerful causal engine could be achieved by synthesising more
heterogeneous causes, mapping mechanisms or noise distribution and pollution
schemes. In a nutshell, RCC's synthesising process should be guided by our
prior knowledge about what kind of data we will typically have to analyze.
This relates in spirit to the Bayesian causal inference strategy proposed by
\citet{Stegle10}, but avoids the need  of expensive, approximate inference at
test time.

\subsection{ChaLearn's ``Fast Causation Coefficient'' challenge}

The data of the ChaLearn's \emph{Fast Causation Coefficient} challenge
\citep{Codalab14} consists on $20,000$ training pairs, $4,050$ validation
pairs, and $4,050$ testing pairs. While the training data was available to the
participants at all times during the competition, both validation and test sets
were stored and predicted remotely on CodaLab's servers.

According to the organizers, the challenge data was generated as follows. Each
causal sample was drawn from the distribution of two random variables $X_i$ and
$Y_i$, that can either be independent, related through a common unobserved
cause, or causing each other.  The data included both heterogeneous real-world
(18\% of the data) and artificial (82\% of the data) causal pairs. The
artificial data was generated using nonlinear models of the form
$Y=f(X,\epsilon)$, $X=g(Y,\epsilon)$, and different noise combination
strategies.  The task of the participants was to use this data to derive a
``causation coefficient'' that would output ``$+1$'' whenever $X \rightarrow
Y$, ``-1'' whenever $Y \rightarrow X$, and zero otherwise. The performance of
each participant was measured using the ``bidirectional AUC'', that is, the
average of the Area Under the Curve (AUC) obtained on the two binary
classification problems ``$X \rightarrow Y$ vs all'' and ``$Y \rightarrow X$ vs
all''.

Our strategy during the competition was as follows. First, we standardized each
variable from each sample to have zero mean and unit variance. Second, we
enlarged the dataset by duplicating each pair by swapping $X$ and $Y$ and its
label accordingly.
Third, we trained a Gradient Boosting Classifier (GBC), with hyper-parameters
chosen via a 4-fold cross validation, on the featurizations \eqref{eq:ourfeatz}
of the training data. In particular, we built two separate classifiers: a first
one to distinguish between causal and non-causal pairs (i.e., $X-Y$ vs
$\{X\rightarrow Y, X\leftarrow Y\}$), and a second one to distinguish between
the two possible causal directions on the causal pairs (i.e., $X\rightarrow Y$
vs $X\leftarrow Y$). The final causation coefficient for a given sample $S_i$
was computed as $$\mathrm{score}(S_i) = p_1(S_i)\cdot(2\cdot p_2(S_i)-1),$$
where $p_1(x)$ and $p_2(x)$ are the class probabilities output by the first and the
second GBCs, respectively. We found it easier to distinguish between causal and
non-causal pairs than to infer the correct direction on the causal pairs.

RCC ranked third in the ChaLearn's ``Fast Causation Coefficient Challenge''
competition, and was awarded the prize to the fastest running code
\citep{Codalab14}.
At the time of the competition, we obtained a bidirectional AUC of $0.732843$
on the test pairs in two minutes of test-time \cite{Codalab14}.
On the other hand, the winning entry of the competition made use of
hand-engineered features, took a test-time of 30 minutes, and achieved a
bidirectional AUC of $0.826413$.
However, some of the hand-engineered features used by this winning entry are of
dubious importance for causal inference, but probably improved its score 
because of the existence of biases in  the challenge data. Examples of these
features are ``unique number of samples from $X$'', ``variance of $Y$'' or
``maximum value of $X$''. 

Interestingly, the performance of IGCI on the $20,000$ training pairs is barely
better than random guessing. The computational complexity of the additive noise
model (usually implemented as two Gaussian Process regressions followed by two
kernel-based independence tests) made it unfeasible to compare it on this
dataset. 

\section{Conclusions and future Work}\label{sec:future}
We have proposed to \emph{learn how to perform causal inference} between pairs
of random variables from observational data, by posing the task as as a
supervised learning problem. In particular, we have introduced an effective and
efficient featurization of probability distributions, based on kernel mean
embeddings and random Fourier features. Our numerical simulations support the
conjecture that patterns revealing causal relationships can be learnt from
data.

We would like to mention three exciting research directions.  First, the use of
the hereby proposed ideas to learn other randomized statistics, such as
measures of dependence \citep{Lopez-Paz13}. Second, the extension of RCC to
operate not on pairs, but sets of random variables, and eventually reconstruct
causal DAGs from multivariate data. Finally, the adaption of the distributional
learning theory of \citet{Szabo14} to analyze our randomized,
classification setting.

\small
\newpage
\bibliography{jmlr_rcc}
\end{document}